\documentclass{article}

\usepackage{microtype}
\usepackage{graphicx}
\usepackage{subfigure}
\usepackage{booktabs} % for professional tables
\usepackage{amsfonts}       % blackboard math symbols
\usepackage{xurl}
\usepackage{nicefrac}       % compact symbols for 1/2, etc.
\usepackage[utf8]{inputenc} % allow utf-8 input
\usepackage[T1]{fontenc}    % use 8-bit T1 fonts
\usepackage{wrapfig}
\graphicspath{ {./Figs/} }

\usepackage{hyperref}

\usepackage[accepted]{icml2021}

\icmltitlerunning{SCC: an Efficient Deep RL Agent Mastering the Game of StarCraft II}

\begin{document}

\twocolumn[
\icmltitle{SCC: an Efficient Deep Reinforcement Learning Agent \\
          Mastering the Game of StarCraft II}



\icmlsetsymbol{equal}{*}

\begin{icmlauthorlist}
\icmlauthor{Xiangjun Wang}{equal,ins}
\icmlauthor{Junxiao Song}{equal,ins}
\icmlauthor{Penghui Qi}{equal,ins}
\icmlauthor{Peng Peng}{ins}
\icmlauthor{Zhenkun Tang}{ins}
\icmlauthor{Wei Zhang}{ins}
\icmlauthor{Weimin Li}{ins}
\icmlauthor{Xiongjun Pi}{ins}
\icmlauthor{Jujie He}{ins}
\icmlauthor{Chao Gao}{ins}
\icmlauthor{Haitao Long}{ins}
\icmlauthor{Quan Yuan}{ins}
\end{icmlauthorlist}

\icmlaffiliation{ins}{inspir.ai, Beijing, China}

\icmlcorrespondingauthor{Xiangjun Wang}{xj@inspirai.com}

\icmlkeywords{Reinforcement Learning, ICML}

\vskip 0.3in
]



\printAffiliationsAndNotice{\icmlEqualContribution} % otherwise use the standard text.

\begin{abstract}
  AlphaStar, the AI that reaches GrandMaster level in StarCraft II, is a remarkable milestone demonstrating what deep reinforcement learning can achieve in complex Real-Time Strategy (RTS) games. However, the complexities of the game, algorithms and systems, and especially the tremendous amount of computation needed are big obstacles for the community to conduct further research in this direction. We propose a deep reinforcement learning agent, StarCraft Commander (SCC). With order of magnitude less computation, it demonstrates top human performance defeating GrandMaster players in test matches and top professional players in a live event. Moreover, it shows strong robustness to various human strategies and discovers novel strategies unseen from human plays. In this paper, we'll share the key insights and optimizations on efficient imitation learning and reinforcement learning for StarCraft II full game.
\end{abstract}

\section{Introduction}
Games as research platforms have fueled a lot of recent advances in reinforcement learning research. The success of Atari \cite{mnih2013playing}, AlphaGo \cite{AlphaGo}, OpenAI Five \cite{dota} and AlphaStar \cite{alphastar} have demonstrated the remarkable results deep reinforcement learning can achieve in various game environments.

As the game complexity increases, those advances come with extremely large computational overhead. For example, in order to train OpenAI Five that reaches Dota 2 professional level, it utilized thousands of GPUs over multiple months \cite{dota}. AlphaStar also trained on hundreds of TPUs for months \cite{alphastar}.

StarCraft, one of the most popular and complex Real-Time Strategy (RTS) games, is considered as one of the grand challenges for reinforcement learning. The reinforcement learning algorithms need to make real-time decisions from combinatorial action spaces, under partially observable information, plan over thousands of decision makings, and deal with a large space of cyclic and counter strategies. Competing with human players is especially challenging because humans excel at reacting to game plays and exploiting opponents' weaknesses.

In this paper, we propose StarCraft Commander (SCC). Similar to AlphaStar, it comprises two training stages, starting with imitation learning, followed by league style reinforcement learning. We'll describe what it takes to train a reinforcement learning agent to play at top human performance with constrained compute resources, as well as the analysis and key insights of model training and model behaviors.

First, we conduct extensive neural network architecture experiments to squeeze the performance gain while reducing memory footprint. For example, reducing the input minimap size from $128\times128$ to $64\times64$ reduces the sample data size almost by half with almost identical performance in the supervised learning stage. We also observed additional performance improvements with various techniques such as group transformer, attention based pooling, conditioned concat attention, etc.

Second, we evaluated the effect of data size and quality for imitation learning. To our surprise, we were able to get most of the performance using only a small number of replays (4,638) compared to the full dataset (105,034 replays). The additional performance gain of large dataset only comes with large batch size. The best result can be obtained with large dataset training with large batch size and fine tuning with small dataset of high quality replays. In the end, the supervised learning model can beat the built-in elite bot with 97\% win rate consistently. 

Third, during the reinforcement learning stage, due to game theoretic design of StarCraft, there exist vast spaces of strategies and cyclic counter strategies. It's crucial to have both strength and diversity so that agents are robust and invulnerable to various counter strategies, which are also the main drivers behind the need for large computational resources. We propose agent branching for efficient training of main agents and exploiters. Even though league training was restricted on a single map and race, the agents exhibit strong generalization playing against other races, on other maps, including unseen ones.

Lastly, SCC was evaluated in test matches with players at different levels. We also held a live match event against professional players. SCC won all the matches against players from GrandMaster to top professionals. According to the feedback from those players, SCC not only learned to play in a way similar to how humans play, but also discovered new strategies that are rare among human games.

% \vspace{-1mm}
\section{Related Work}
In this section, we briefly review early work for StarCraft AI and describe AlphaStar algorithms.

StarCraft is a popular real time strategy game involving strategy planning, balance of economy and micromanagement, game theoretic challenge. Those combined challenges make StarCraft an appealing platform for AI research. StarCraft: Brood War has an active competitive AI research community since 2010 \cite{ontanon2013survey,weber2010aiide}, where most bots are built with heuristic rules together with search methods \cite{churchill2013portfolio, churchill2017analysis}. There has been some work using reinforcement learning for mini-games and micromanagement \cite{peng2017multiagent,vinyals2017starcraft,zambaldi2018relational,foerster2017counterfactual,usunier2016episodic}. Most recently, reinforcement learning was used to play the full game, combined with hand-crafted rules \cite{sun2018tstarbots,lee2018modular,pang2019reinforcement}. Even though some of the bots successfully beat the game built-in AI \cite{sun2018tstarbots,pang2019reinforcement}, none of the early work reached competitive human level.

AlphaStar is the first end-to-end learning algorithm for StarCraft II full game that reaches GrandMaster level. It uses imitation learning to learn initial policy from human replay data, which not only provides a strong initialization for reinforcement learning, more importantly, it learns diverse sets of human strategies that are extremely hard for reinforcement learning to learn from scratch. It also uses statistic $z$ to encode build orders and build units for guiding play strategy. In addition to self-play, league training is adopted for multi-agent reinforcement learning. The league consists of three distinct types of agents for each race, main agent, main exploiter and league exploiter. First, the main agents utilize a prioritized fictitious self-play (PFSP) mechanism that adapts the mixture probabilities proportionally to the win rate of each opponent against the agent, to dynamically focus more on the difficult opponents. Second, main exploiters play only against current main agents to find weaknesses in main agents. Third, league exploiters use a similar PFSP mechanism against all agents to find global blind spots in the league. They work together to ensure the main agents improve strength and robustness when competing with human players.

AlphaStar introduced the first version with high level ideas described on a blog post \cite{alphastarblog}. It specialized in race Protoss and was evaluated against two professional players. A revised version was published later \cite{alphastar}. The later version changed the mechanism of league training to be more generic, utilized statstic $z$ to encode build order and units, and trained all three races with constrained actions per minute (APM) and camera interface setting. On the infrastructure side, a total of 12 separate training agents are instantiated with four for each race, and for every training agent, it runs 16,000 concurrent StarCraft II matches to collect samples and the learner processes about 50,000 agent steps per second. It was evaluated on the official online matching system Battle.net and rated as the top level (GrandMaster) on the European server.

TStarBot-X \cite{han2020tstarbot} is a recent attempt to reimplement AlphaStar, with specialization in race Zerg. It encountered difficulties reimplementing AlphaStar's imitation learning and league training strategy. To overcome those issues, it introduced importance sampling in imitation learning, rule-guided policy search and new agent roles in league training. Those methods helped efficiency and exploration but it had to incorporate multiple hand-crafted rules such as rule-guided policy search and curated datasets for 6 fine-tuned supervised model. In the end, the human evaluation showed comparable performance with two human Master players whose expertise are not race Zerg. 

\section{StarCraft Commander (SCC)}
To the best of our knowledge, SCC is the first learning-based agent that reaches top human professional level after AlphaStar, while using order of magnitude less computation and without using any hand-crafted rules. SCC was developed around the time the first version of AlphaStar was published \cite{alphastarblog} and adopted the main ideas from it. SCC interacts with the game of StarCraft II (getting observations and sending actions) using the s2client protocol \cite{s2client-proto} and the PySC2 environment \cite{pysc2}, provided by Blizzard and DeepMind respectively.

Without knowing all the AlphaStar algorithms details at the time, we independently experimented with network architecture, imitation and reinforcement learning training mechanisms. Given our computational constraint, we did extensive optimizations to squeeze the efficiencies out of each stage of learning. Some of the major differences with AlphaStar are highlighted below:
\begin{itemize}
    \item Game setup is similar to the first version of AlphaStar \cite{alphastarblog} except that SCC was trained on race Terran, one of the more challenging races for AI to learn. Although the algorithm is generic to any race, we focus on one race to save computation and evaluation efforts.
    \item AlphaStar uses 971,000 replays played on StarCraft II versions 4.8.2 to 4.8.6. SCC only has access to a much smaller dataset, with 105,034 replays on versions 4.10.0 to 4.11.2. 
    \item SCC's network architecture is similar to AlphaStar, but more memory efficient, with additional optimizations such as group transformer, attention-based pooling, etc. SCC's final reinforcement learning model has 49M parameters (139M for AlphaStar) \cite{alphastar}. 
    \item Each agent of SCC is trained using Proximal Policy Optimization (PPO) \cite{PPO} in the reinforcement learning stage, with further modifications to utilize asynchronous sampling and large scale distributed training.
    \item SCC also uses league training consisting of main agent, main exploiter and league exploiter, but with multiple main agents for more diversity. The agent branching approach effectively improved learning efficiency of main agents and main exploiters, resulting in strong and robust pool of agents.
    \item SCC uses much fewer samples for learning. For each agent, 1000 StarCraft environments are used to collect samples (16,000 for AlphaStar) and the learner consumes about 800 agent steps per second (50,000 for AlphaStar). In total, each agent experienced 30 years of real-time StarCraft play (200 game years for AlphaStar) \cite{alphastarblog,alphastar}.
    \item Since we don't have access to official battle.net, we invited human players of different competitive levels (from Diamond to GrandMaster) to each play five matches with SCC at different strength levels and SCC won all of them. In the end, we held a live event to play two sets of best of three matches with two top professional players, SCC won both games with 2:0.
\end{itemize}

In the following sections, we describe the differences mentioned above in detail, including network architecture, imitation learning, reinforcement learning and evaluations. For brevity, the training platform is described in detail in Appendix \ref{appendix-platform}.

\section{Network Architecture}
With the same input and output interfaces provided by the StarCraft II game engine, the overall network architecture of SCC is similar to that of AlphaStar. From input to output, the model architecture can be roughly divided into three parts, i.e., input encoders, aggregator, and action decoders. The input encoders take in three groups of observations, i.e., scalar features, spatial features and sets of units features, process them separately and yield encoded feature vectors accordingly. The encoded feature vectors are fed into the aggregator, which concatenates all of them and feeds the resulting vector into a residual LSTM block. Then, the LSTM output is fed into the action decoder, which decides the final action based on the outputs of its six heads. For more detailed network architecture, please refer to Fig. \ref{fig:network} in Appendix \ref{appendix-network}.

We also would like to further highlight some key designs, which are the main contributors for improved computational efficiency and expressive power of network architecture, enabling learning a strong model using fewer than 5,000 replays during imitation learning.

\begin{itemize}
    \item Network pruning: we use a spatial input size of 64x64 which reduces around half of the per-sample data size compared to 128×128 for AlphaStar. In addition, instead of the very deep 16 blocks of residual MLP used in AlphaStar, a simple fully connected layer is chosen for the selected action head in the action decoder. With these simplifications, almost identical performance is observed in supervised learning.

    \item Group transformer: among all the observations, the set of units is most informative. In AlphaStar, all units are put together and processed by a transformer. We instead divided them into three groups, my units, enemy units and neutral units. Since they are naturally heterogeneous, separate learning allows more degrees of freedom. Specifically, for each group of units, one multi-head self-attention block is applied to the units within the group, and two multi-head cross attention blocks are applied to units between groups. The outputs of the three blocks are concatenated together as the ﬁnal encoded unit features.

    \item Attention-based pooling: the unit features encoded by multi-head attention blocks are set of vectors. A simple average-pooling is applied in AlphaStar to reduce them into a single vector. We propose attention-based pooling, in which trainable weight vectors are created as the queries, and the unit features are reduced by a learned weighted averaging.

    \item Conditional structures: to deal with the combinatorial action space, AlphaStar uses an additive auto-regressive action structure. Instead of the additive operator, we adopt the structure of concatenation followed by a fully connected layer, to provide full ﬂexibility for the network to learn a better conditional relationship. In addition, different selected actions may have totally different criteria when selecting the target, e.g. the repair action tends to target damaged alliance units and the attack action tends to target nearby enemy units. In light of this, we propose the conditioned concat-attention for target selection heads, in order to learn different selection functions for different selected actions.
\end{itemize}

\section{Imitation Learning}
SCC was initially trained by imitating games sampled from publicly available human replays. Supervised learning not only provides good initialization for the following reinforcement learning stage but also diverse human strategies that are extremely difficult to explore from scratch. Also, supervised learning makes it easy to evaluate policy architecture. A good network architecture for supervised learning is also likely to be good for reinforcement learning. The current network architecture of SCC is the result of extensive experiment studies. Details can be found in Appendix \ref{appendix-network}. In this section, we describe the settings applied in the supervised learning stage and also provide ablation study results regarding training configurations. Evaluation results of the model obtained from supervised learning will also be presented.

\subsection{Dataset and Setting}
We used a dataset of 105,034 replays played on StarCraft II versions 4.10.0 to 4.11.2 by players with MMR scores greater than 4,300. Instructions for downloading replays can be found at \url{https://github.com/Blizzard/s2client-proto}. Note that since we only trained an agent for race Terran, the dataset only consists of replays of three race match-ups, i.e., Terran versus Terran, Terran versus Protoss and Terran versus Zerg, with 34,487, 39,391 and 31,156 replays respectively.

\subsection{Training Analysis}
\begin{figure*}[ht]
    \centering
    \includegraphics[width=0.96\textwidth]{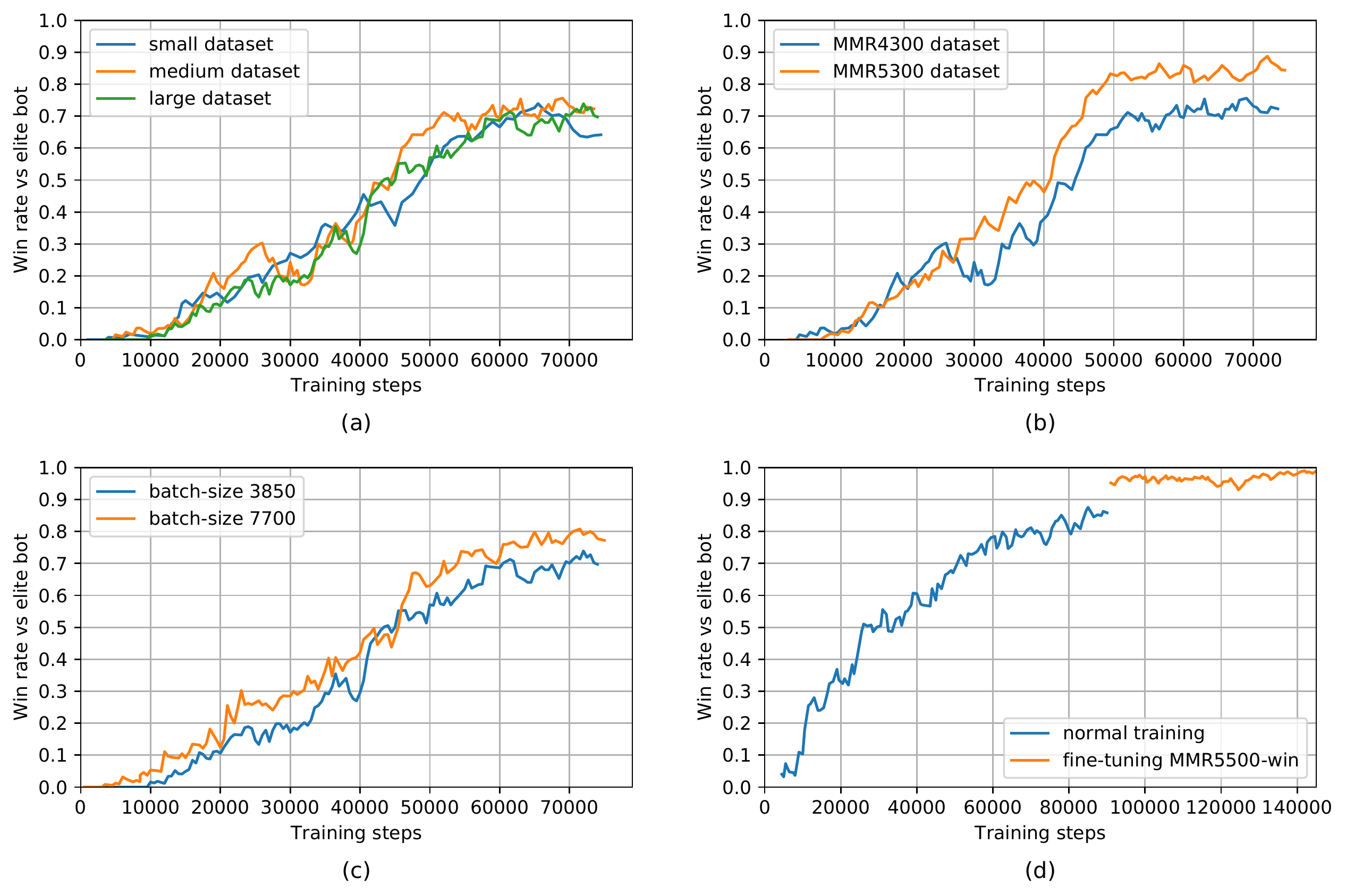}
    \vskip -0.1in
    \caption{Win rates versus the built-in elite bot during the supervised training progress of different trials. \textbf{(a)} Training runs using three datasets of different sizes. The small, medium and large datasets all consist of replays with MMR scores greater than 4,300 and with 4,638, 15,579 and 105,034 replays, respectively. \textbf{(b)} Training runs using two datasets of different MMR scores. The MMR4300 dataset is just the medium dataset, which consists of 15,579 replays and the MMR5300 dataset consists of 17,173 replays. \textbf{(c)} Training runs using the same large dataset but with different training batch sizes. \textbf{(d)} The final supervised training starting with a normal training stage and followed by a fine-tuning stage.}
    \label{fig:training_analysis}
    \vskip -0.05in
\end{figure*}

\paragraph{Dataset size and quality}
In supervised learning, the size and the quality of the dataset matter. We conducted several ablation experiments regarding the impact of the datasets. First, we constructed three datasets of different sizes, with 4,638, 15,579 and 105,034 replays respectively. In the following, we name them as small, medium and large dataset accordingly. The large dataset is the full dataset, the small dataset and medium dataset are subsets of replays sampled from the large dataset. We trained supervised models using these three datasets, with batch-size 3,850. The learning rate was set to 1e-3 before 40,000 training steps and was decayed to 1e-4 after that. During the training progress, we evaluated the models against the built-in elite bot, and the results are shown in Fig. \ref{fig:training_analysis} (a). Surprisingly, we observe that the three resulting models perform quite similar, with peak win rates all slightly above 70\%. Given the huge complexity of the StarCraft II full game, it is really amazing that the model trained using a small dataset with fewer than 5,000 replays can already play reasonably well. One hypothesis is the efficient network architecture enables efficient learning. Our models during early version of development could not achieve similar performance even with more data. It suggests design of model architecture really matters in this case. Although we do observe the performance of the model using small dataset decreases quickly after the peak, indicating a possibility of overfitting.

% % \begin{figure}[htp]
% %     \centering
% %     \includegraphics[width=12cm]{Figs/dataset_size.pdf}
% %     \caption{Win rates versus the built-in elite bot during the training progress using three datasets of different sizes. The small, medium and large datasets all consist of replays with MMR scores greater than 4300 and with 4638, 15579 and 105034 replays respectively.}
% %     \label{fig:dataset_size}% 
% % \end{figure}

We also analyzed the impact of the quality of the dataset, in terms of the MMR score. We filtered out all replays with MMR score greater than 5,300 from the large dataset, and obtained a dataset of size 17,173, which is close to the size of the medium dataset. We call it the MMR5300 dataset. Then we trained a supervised model using the MMR5300 dataset, with all other settings same as above. The evaluation results of the resulting model and that of the medium dataset (MMR4300 dataset) are plotted together in Fig. \ref{fig:training_analysis} (b). It is obvious that the performance of the model trained with the MMR5300 dataset is almost always above that of the MMR4300 dataset and the peak performance is around 85\%, about 10\% higher than the MMR4300 dataset. 

% % \begin{figure}[htp]
% %     \centering
% %     \includegraphics[width=12cm]{Figs/dataset_quality.pdf}
% %     \caption{Win rates versus the built-in elite bot during the training progress using two datasets of different MMR scores. The MMR4300 dataset is just the medium dataset, which consists of 15579 replays and the MMR5300 dataset consists of 17173 replays.}
% %     \label{fig:dataset_quality}% 
% % \end{figure}

\paragraph{Batch size}
In the previous paragraph, we have observed that larger dataset did not lead to better performance. One intuitive idea is that larger dataset may be more difficult to fit, in order to get performance gain, we may need to use a larger batch size. We take the performance of the large dataset in Fig. \ref{fig:training_analysis} (a) as the baseline, which was trained with batch size 3,850. With all other training settings fixed, we doubled the batch size to 7,700 and ran another training. The evaluation results during the training progress are shown in Fig. \ref{fig:training_analysis} (c). We can observe that with batch size doubled, the performance is indeed improved, in terms of both the learning speed and the peak win rate.

\paragraph{Final supervised training}
Based on the experiment results above, we trained the final supervised model with large dataset first, and then fine tuned with a small but high quality dataset for additional gain. Specifically, we first trained the model using the full dataset with learning rate 1e-3 and batch size 15,400, and then fine tuned the model with the MMR5500-win dataset and learning rate decayed to 1e-5. MMR5500-win dataset is of size 3,458 and contains only the winning replays with MMR score above 5,500. The evaluation performance during this final supervised training progress is shown in Fig. \ref{fig:training_analysis} (d). At the late fine-tuning stage, win rate against built-in elite bot exceeds 97\% consistently.

% % \begin{figure}[htp]
% %     \centering
% %     \includegraphics[width=12cm]{Figs/sl-final.pdf}
% %     \caption{Win rates versus the built-in elite bot during the final supervised training.}
% %     \label{fig:sl-final}% 
% % \end{figure}

\subsection{Evaluation}
\paragraph{Against built-in AIs} 
The supervised learning agent was evaluated against built-in AIs of different difficulties, on different ladder maps, and in Terran versus Terran, Protoss and Zerg games. The results are shown in Table \ref{sl-eval-table}. The built-in elite bot and the CheatInsane bot are used as the opponents, which are the strongest built-in non-cheating AI and the strongest built-in AI that cheats in vision and resources, respectively. Results on three ladder maps, i.e., Triton, KairosJunction and Catalyst, are reported. Note that our training data does not contain Catalyst, which is a ladder map in earlier versions of StarCraft. From the results, we can see that among the three opponent races, the agent achieves the highest win rates in games of Terran versus Terran. Among the three maps, the two seen during training give higher win rates as expected, but surprisingly the agent also plays relatively well on the unseen map Catalyst, which exhibits the generalization capability of the supervised model to different maps.

\begin{table*}[ht]
  \caption{Win rates (out of 100 matches) of the final supervised model versus the built-in elite bot and the CheatInsane bot, on three different maps: Triton, KairosJunction and Catalyst. Catalyst is an old map before 2018 season 2 not included in the dataset and thus not seen during training. The elite bot is the strongest built-in non-cheating AI, and the CheatInsane bot is the strongest built-in AI that cheats in vision and resources, i.e., having full vision on the whole map and resource harvest boosting.}
  \label{sl-eval-table}
  \centering
  \begin{tabular}{lcccccc}
    \toprule
    & \multicolumn{2}{c}{TvT}  & \multicolumn{2}{c}{TvP}  & \multicolumn{2}{c}{TvZ}   \\
    \cmidrule(r){2-3}  \cmidrule(r){4-5}  \cmidrule(r){6-7}
    Map & Elite & CheatInsane & Elite & CheatInsane & Elite & CheatInsane \\
    \midrule
    Triton & 0.97 & 0.29 & 0.91 & 0.10 & 0.95 & 0.11     \\
    KairosJunction & 0.94 & 0.38 & 0.90 & 0.21 & 0.93 & 0.23     \\
    Catalyst & 0.90 & 0.19 & 0.87 & 0.08 & 0.91 & 0.09     \\
    \bottomrule
  \end{tabular}
\end{table*}

\begin{figure*}[ht]
    \centering
    \includegraphics[width=0.85\textwidth]{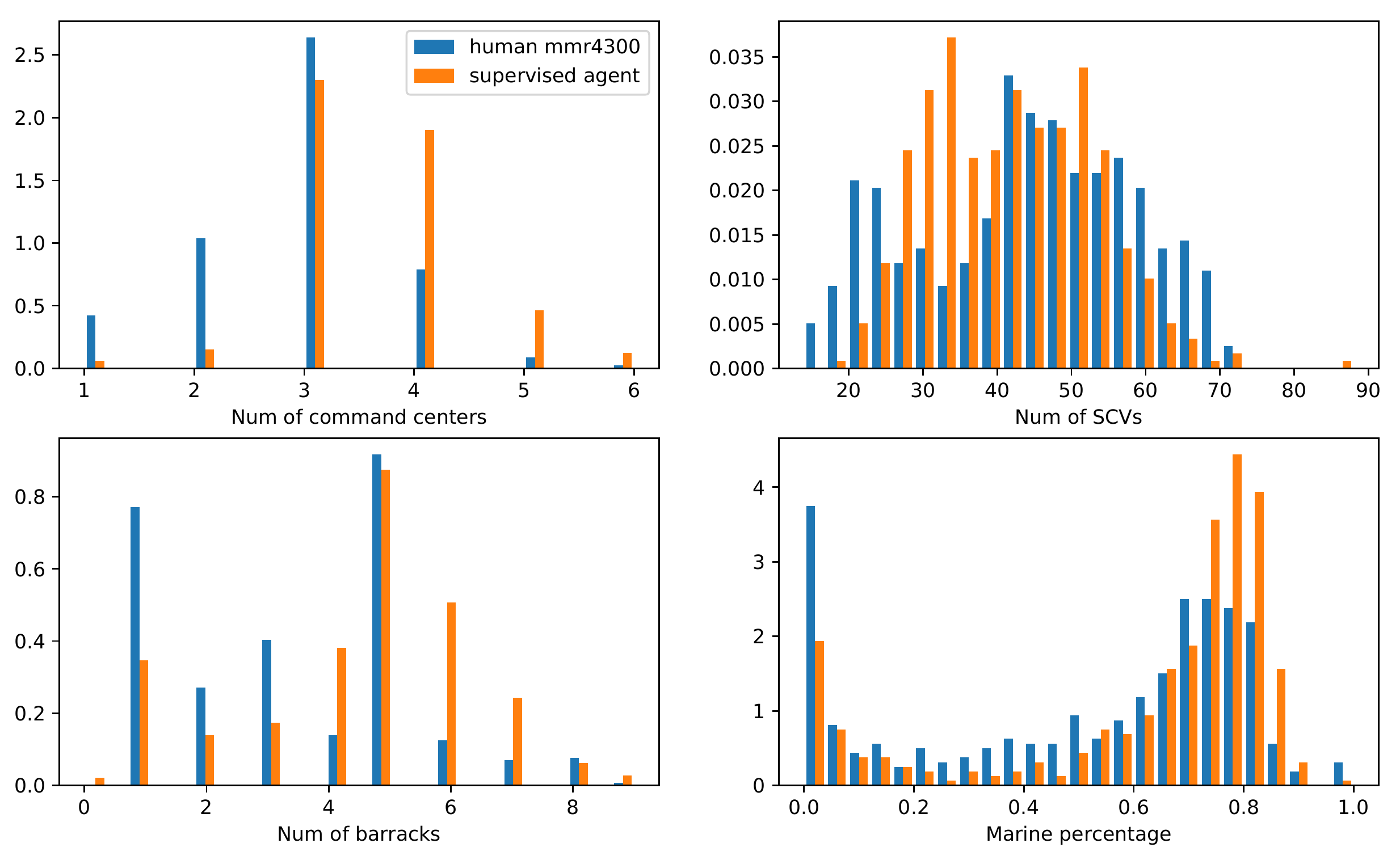}
    \caption{Comparison of some statistic distributions of human data and final supervised agent.}
    \label{fig:human-sl-dist}
\end{figure*}

\paragraph{Comparison with human policy}
To compare the policy learned by supervised learning with that of humans, we let the final supervised agent play against itself on map Triton for 400 games and calculated four indicative statistics of playing strategies, i.e., number of command centers, number of SCVs, number of barracks, and the percentage of Marine unit in all trained units, for each of the games. Similarly, we randomly picked 400 Terran versus Terran human replays on the same map with MMR scores above 4300 and calculated the same four statistics for each of the replays. The distribution densities of the four statistics of both the supervised agent and the humans are shown in Fig. \ref{fig:human-sl-dist}. From the figure, we can observe similar distribution shapes for all statistics, e.g., most of the times three command centers and five barracks are built for both the supervised agent and humans, which indicate that the supervised agent indeed plays a similar set of strategies to humans.

\section{Reinforcement Learning}
SCC was trained in the setting of Terran versus Terran on a single ladder map Triton at the reinforcement learning stage. The final supervised agent was used as the initialization for reinforcement learning stage. In the following, we will describe the training procedures that have been used during the course of training. Detailed description of the training platform is in Appendix \ref{appendix-platform}. 

\subsection{Learning Algorithm}
Starting from the policy provided by the supervised training stage, we apply the Proximal Policy Optimization (PPO) \cite{PPO} to further improve the performance of SCC based on agent versus agent games. To improve the sampling and training efficiency, we modified the original PPO to utilize the efficiency of asynchronous sampling and large scale distributed training. In addition to the standard PPO loss, we also added extra loss terms, i.e., a standard entropy regularization loss, and a KL divergence loss with respect to the final supervised policy to prevent RL policy deviating from humans \cite{alphastar}. Thus the overall loss used in reinforcement training is as follows:
\[L_{RL}=L_{PPO} + L_{entropy} + L_{KL}.\]

\paragraph{Rewards}
We found the statistic $z$ used in AlphaStar a very useful approach to automate the training process and aid in exploration and diversity throughout learning, therefore we also adopted a similar approach. Besides the sparse win/loss reward, there is a second type of reward based on strategy statistic $z$. The strategy statistic $z$ are extracted from human replays which encode constructed buildings and units present during a game. For the reinforcement learning stage, we collected only the $z$ from Terran versus Terran replays on Triton map. At the start of each game, we randomly sample a $z$, and at each following agent step, the rewards are calculated as the distances between each part of the sampled statistic $z$ and the actually executed statistic. It encourages the agent to follow different human strategies, and thus help to preserve diversity. Separate value functions and advantages are computed for each reward, and the generalized advantage estimation (GAE) \cite{GAE} is applied to balance the trade-oﬀ between the bias and variance of the advantage estimation.

\paragraph{Shared policy and value network with warm-up}
Although the policy network is initialized with the supervised model, the value network is missing. We observed that at the beginning of reinforcement learning, the value output is very noisy, resulting in noisy calculated advantage and unstable training. To mitigate the problem, we freeze the policy network and update only the value network at the first 50 training steps for warm up. In addition, the policy and value networks share the same network architecture and weights before the LSTM output. With the shared architecture, the value network only needs to learn a mapping from the encoded LSTM output to the predicted value, instead of learning a whole mapping from the raw observations, which may also ease the warm-up of the value network. Additionally, the shared architecture also reduces the number of weights, i.e., the model size, so that we can use larger batch size and accelerate the training, which is critical especially when we have only limited memory and computational resources. 

\subsection{Agent League}
We found the general self-play approach insufficient to address the game theoretic challenge of StarCraft, and adopted league training approach similar to AlphaStar, using a combination of main agent, main exploiter and league exploiter.
%To ensure robustness of the agent and avoid strategy cycles, we follow the league training approach used in AlphaStar but with a few changes. The league consists of three different types of agent, namely the main agent, the main exploiter and the league exploiter, which differ in their mechanism of selecting opponents and also the rewards they used. First, the main agent plays against the whole league, i.e., all the players in the league, including itself, and the prioritized fictitious self-play (PFSP) mechanism is applied. For the main agent, among the two types of rewards mentioned in the previous subsection, the win/loss reward is always active, while the rewards based on strategy statistic $z$ are active independently with probability 25\% for each part. With this reward setting, the main agent can play various strategies when different statistic $z$ are specified. Second, the main exploiter plays against only the current main agent and it aims to exploit the weakness of the main agent. Third, the league exploiter plays against the whole league except the current main agent. It is used to identify systemic weakness of the league. Of course, by finding these weaknesses, the goal is to address them by playing against the exploiters.  

\paragraph{Agent branching}
During the league training progress of SCC, we found that as the quality and quantity of agents in the league increases, it takes longer to train an exploiter starting from the supervised agent. After analyzing the behavior of main agent with different statistic $z$ specified, we observed that the main agent can play partially according to the specified $z$ but not perfectly, likely due to the limited data for a specific $z$, since there are hundreds of them. And when playing against the unconditional main agent, different $z$ lead to different win rates range between 35\% to 65\%.

Based on these observations, we propose an approach to train agents more efficiently with specified strategies. Specifically, instead of initializing with the supervised agent and training with sparse win/loss reward, we propose to initialize the new agent with the current main agent and train it with not only the sparse win/loss reward, but also the dense reward based on a specific $z$ or a set of $z$. The specified $z$ could be the one with the highest win rate against the unconditional main agent if we want to train an exploiter quickly, or any specific strategy as needed. With this approach, the new agents branch off the main agent and move toward different directions, thus we call the approach agent branching.

In addition to reducing training time, we find the agent branching approach very helpful for agent robustness to human strategies. For example, reaper rush is an early aggressive strategy commonly used by human players and it's very effective if executed well. We find it is really difficult for the exploiters to discover and execute similar strategies very well if only sparse win/loss reward is used. Using agent branching, exploiters can be guided during learning to specialize in a strategy, resulting in a much stronger exploiter. Thus the main agent learns to cope with the reaper rush strategy very well, even when played against human players.

\paragraph{Multiple main agents}
With reduced computational resources required for the exploiters, we explored the idea of training multiple main agents. During the course of SCC league training, three main agents were trained, each starting at a different stage. Among them, two were initialized with the supervised model, and one was trained using the agent branching approach. With multiple main agents, we found them play differently even for those two initialized with the same supervised model. Agents initiated at different stage of league training could evolve differently due to the dynamics of league training. They provide more diversity to the league, thus making the trained agents more robust.

\subsection{Evaluation}

\begin{figure}[ht]
% \vskip 0.2in
\begin{center}
\centerline{\includegraphics[width=\columnwidth]{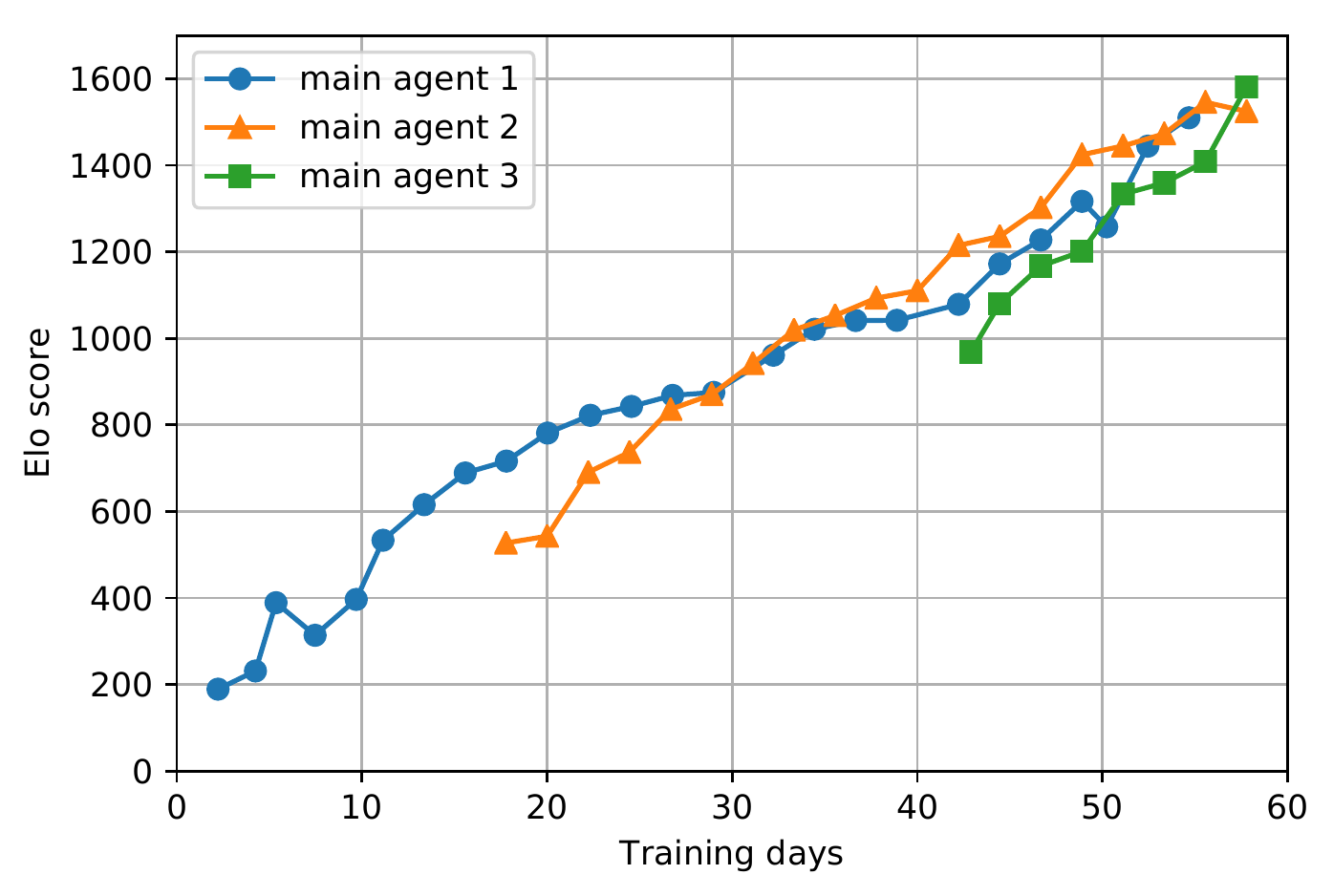}}
\caption{Elo scores of three main agents during the 58 days of training. Each point represents a past checkpoint of the main agents. The Elo score of the final supervised agent is set to 0 as the baseline.}
\label{fig:elo-score}
\end{center}
\vskip -0.3in
\end{figure}

\paragraph{Elo scores of the main agents}
To examine the progression of the league training, we consider the rating metric Elo score. Elo score determines the rating difference between consecutive pairs of models based on the win rates between them. The supervised agent was chosen as the baseline with Elo score set to zero. The Elo score curves of the three main agents over the course of league training are plotted in Fig. \ref{fig:elo-score}. Main agent 1 was the first main agent, initialized with the supervised model, and was trained longest for about 58 days. Main agent 2 was added into the league about 15 days later than main agent 1, and was also initialized with the supervised model. After being added into the league, Main agent 2 played against all existing players in the league, and caught up main agent 1 in Elo score within about 15 days. Main agent 3 was obtained using the agent branching approach based on main agent 1, and was only trained for about 15 days. The Elo scores of all three main agents increase gradually during the course of training, and are above 1,500 at the end.

\paragraph{Generalization of the agent}
The increasing Elo scores exhibit the desired progression of agents, but those are observed in the setting of Terran versus Terran games and on map Triton, under which the agents are trained. It is well known that agents trained by reinforcement learning are prone to overfitting to the training environment. A natural question is whether the final agents can still play against other races and on maps that are not seen during reinforcement training. To answer this question, we perform evaluations against built-in AIs again similar to the evaluation of the supervised agent. The evaluation results are shown in Table \ref{rl-eval-table}. It shows that the final reinforcement learning agent not only can play against untrained races, on unseen maps, but also with dramatically improved win rates. It's an indication that during the league training, the agent indeed learns the essentials to play StarCraft II well in general, instead of just memorizing and overfitting to the training environment.

% \begin{figure}[htp]
%     \centering
%     \includegraphics[width=12cm]{Figs/elo-score.pdf}
%     \caption{Elo scores of three main agents during the 58 days of training. Each point represents a past checkpoint of the main agents. The Elo score of the final supervised agent is set to 0 as the baseline.}
%     \label{fig:elo-score}
% \end{figure}

\begin{table*}[ht]
  \caption{Win rates (out of 100 matches) of the final reinforcement agent versus the built-in elite bot and the CheatInsane bot, on three different maps: Triton, KairosJunction and Catalyst. The agent is trained only on map Triton in the setting of TvT during reinforcement learning.}
  \label{rl-eval-table}
  \centering
  \begin{tabular}{lcccccc}
    \toprule
    & \multicolumn{2}{c}{TvT}  & \multicolumn{2}{c}{TvP}  & \multicolumn{2}{c}{TvZ}   \\
    \cmidrule(r){2-3}  \cmidrule(r){4-5}  \cmidrule(r){6-7}
    Map & Elite & CheatInsane & Elite & CheatInsane & Elite & CheatInsane \\
    \midrule
    Triton & 1.0 & 1.0 & 1.0 & 1.0 & 1.0 & 0.99     \\
    KairosJunction & 1.0 & 1.0 & 1.0 & 1.0 & 1.0 & 0.96     \\
    Catalyst & 1.0 & 0.99 & 1.0 & 1.0 & 1.0 & 0.96     \\
    \bottomrule
  \end{tabular}
\end{table*}

\begin{table*}[ht]
  \caption{Test results of SCC when playing against human players from Diamond level to Grandmaster level.}
  \label{human-eval-table}
  \centering
  \begin{tabular}{lccccc}
    \toprule
    \multicolumn{3}{c}{Human player}  & \multicolumn{2}{c}{SCC}  & Match result   \\
    \cmidrule(r){1-3}  \cmidrule(r){4-5}  
    ID & MMR score & Ladder tier & Training days & Avg estimated Elo score & SCC vs Human \\
    \midrule
    1 & 3800 & Diamond & 23 & 765 & 5 : 0     \\
    2 & 5000 & Master & 30 & 905 & 4 : 1     \\
    3 & 5500 & GrandMaster & 39 & 1070 & 3 : 2     \\
    4 & 5200 & GrandMaster & 47 & 1240 & 5 : 0     \\
    \bottomrule
  \end{tabular}
\end{table*}

\section{Human Evaluation}
Human evaluation is the final benchmark for StarCraft AI. To evaluate SCC without access to official Battle.net, we invited human players of different levels to play against SCC. All SCC versus human matches are conducted in Terran versus Terran games, on the ladder map Triton, and using StarCraft II of version 4.10.0. During the matches, SCC interacts with the StarCraft game engine directly via its raw interface. In real-time evaluation, SCC reacts with a delay between observation and action, due to two reasons. First, after a frame is observed, SCC needs to process the observation as a valid input for the policy network and then perform a forward inference, which takes about 100ms on average. Second, at each step, SCC decides when to observe next with a delay between 2 to 128 game frames, i.e., 90ms to 5.7 seconds, and is on average about 420ms. Due to the delays, SCC is limited in the number of actions per minute (APM), and in practice we found that the agent would not act at the limit, probably because it learns from human replays. During the course of reinforcement training, we observed that the average APM of SCC gradually increased from 250 (with peak APM below 600) to around 400 (with peak APM below 1000), which is comparable to top human players.

\subsection{Test Matches}
We evaluated the performance of SCC at different stages by playing against human players of different levels, ranging from Diamond to Grandmaster on the Korean server. During the test matches, five games were played in total. Since SCC has multiple main agents, a random one was picked for each game. The results of the test matches are shown in Table \ref{human-eval-table}. In general, human players are extremely good and quick in responding and adjusting strategies during multiple consecutive games, which imposes great challenges on robustness of AI. During some of the games, we even intentionally asked human players to play cheese strategies to exploit the weaknesses of the agent, thanks to the diverse sets of exploiters during the league training, SCC reacted really well to those strategies and demonstrated strong robustness.

\subsection{Live Matches}
On June 21, 2020, we held a live match to challenge top professional players of StarCraft II. TIME \cite{TIME:online} and TooDming \cite{TooDming:online} were invited to play against SCC in two best of three (BO3) matches. Both of them are StarCraft II professional champion winners. The final versions of the three main agents of SCC were used in the matches, with estimated average Elo score above 1,500. As a result, SCC won the two matches with two 2:0. Two narrated replays can be viewed at \url{https://youtu.be/yy8hc4CONyU}. During the first match versus TIME, SCC adopted a strategy with massive landed Vikings, which is seldom used by human players. The strategy was also well received in StarCraft II community and some players have started adopting the strategy in their own games. 

Usually it's expected that AI has an advantage in precise unit control compared to human players. However according to the feedback from TIME and TooDming, although there are still rooms for improvement in terms of micromanagement and units control, it's really impressive that SCC learns to optimize for long term economy and macro, with relentless efforts in constant harassing and expansion, which is often observed among the games of top professional players.

% \section{Discussion}

\section{Conclusion}
We present StarCraft Commander (SCC), a deep reinforcement learning agent for StarCraft II that demonstrates top human performance defeating GrandMaster players in test matches and top professional players in live matches. We describe the key ideas and insights for imitation learning and reinforcement learning. With optimized network architecture and training methods, we reduce the computation needed by an order of magnitude compared to AlphaStar.

There are a lot of interesting observations from the experiments. The ability to learn a good policy to play StarCraft II full game from a very small dataset is truly surprising. SCC also demonstrates unexpected generalization capability for unseen maps and races during reinforcement learning. Finally, the fact SCC is able to discover novel and long term strategies that inspire the StarCraft game community is very encouraging. We believe it's a very promising area of research to further optimize the balance of exploration and exploitation of RL policies, for the emergence of more novel strategies.

Even though the experiments are conducted on StarCraft II, the approaches are generic to other tasks. Our goal is to provide insights for the research community to explore large scale deep reinforcement learning problems under resource constraints. As demonstrated through our experiments, exciting opportunities lie ahead in exploring sample efficiency, exploration and generalization.

\clearpage
\bibliography{scc}
\bibliographystyle{icml2021}

% \newpage
\clearpage
\appendix
\section{Network Architecture}\label{appendix-network}
\begin{figure*}[ht]
    \centering
    \includegraphics[width=\textwidth]{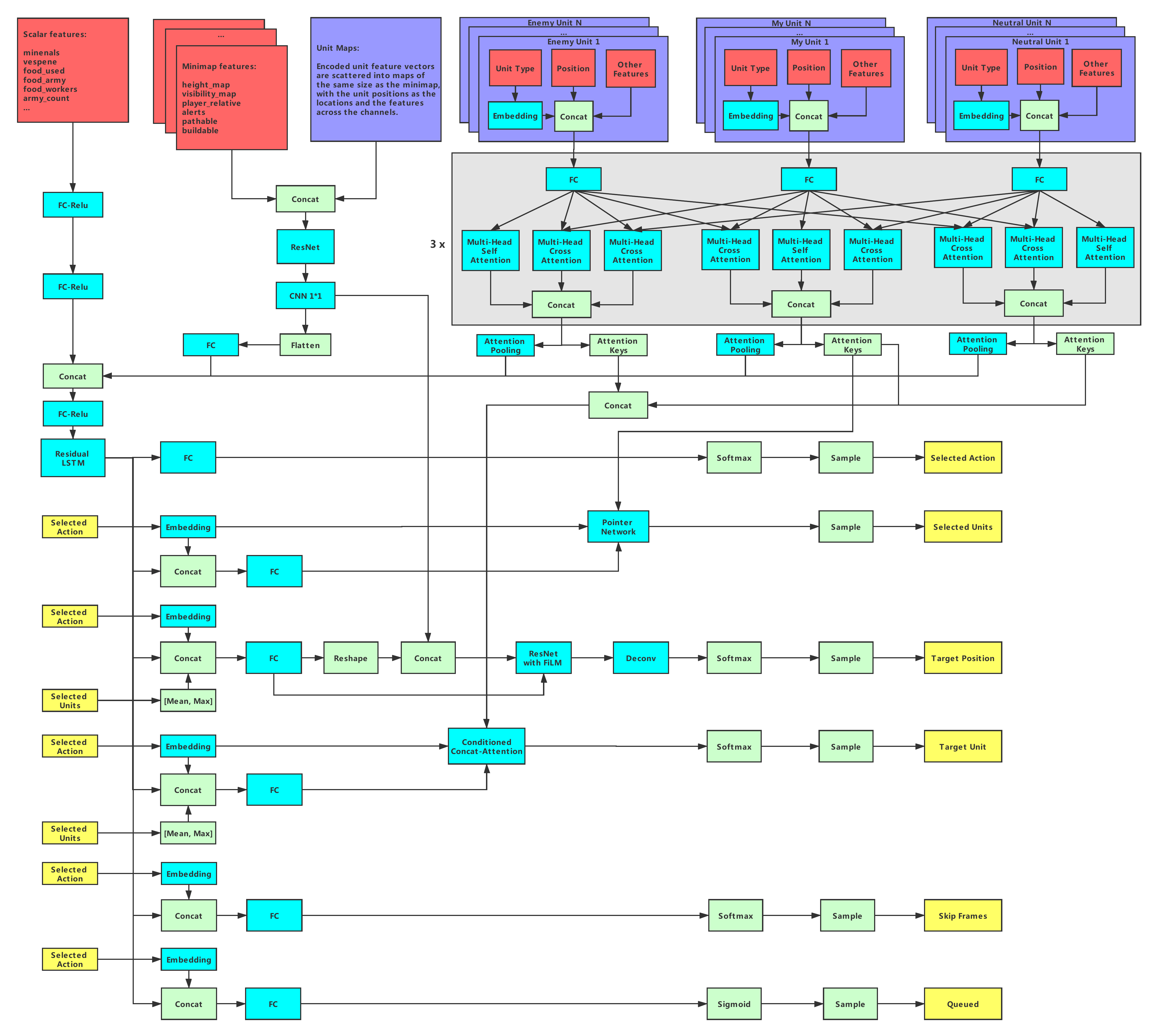}
    \caption{Overview of the policy network architecture of SCC.}
    \label{fig:network}
\end{figure*}

With the complex observation and action structures, the design of the policy network architecture is critical, in order to extract the right information from the raw observation and issue the proper action. The overall architecture of the policy network of SCC is shown in Fig. \ref{fig:network}, which was designed based on the performance in supervised learning. With the same input and output interfaces provided by the game engine, the overall policy structure of SCC is similar to that of AlphaStar, while there are some key components that are different, which will be discussed in detail below.

\subsection{Minimap Size}
As part of the observation, the minimap consists of 6 planes of spatial information, i.e., height, visibility,  player relative, alerts, pathable and buildable, which is directly provided by the raw interface and its size can be specified. AlphaStar uses minimaps of size $128\times128$, as it is the only part of spatial input, it also consists most of the data size. We reduce the minimap size from $128\times128$ to $64\times64$, which reduces the per sample data size by around 49.5\%. We compared the overall performance of the two settings in supervised learning, and observed that the two different minimap sizes lead to almost identical training and evaluation results. As a result, we finally adopt minimap of size $64\times64$, for smaller storage and higher computational efficiency.

\subsection{Group Transformer}
Among all parts of the observations, the set of units is most informative, with each one consists of the unit type, owner, position and many other fields. The whole set of units can be divided into three groups, namely my units, enemy units and neutral units. In AlphaStar, all three groups of units are put together and is then processed by transformer. In our point of view, the three groups of units are naturally separate and processing them separately allows a more comprehensive expression. It will also facilitate the processing later, for example, the selected units head can only select units from my units.

For each group of units, one multi-head self attention block is applied with the unit feature vectors in the same group as the queries, keys and values, and two multi-head cross attention block are applied with the queries being the unit feature vectors in the same group and that of the other two unit groups being the keys and values \cite{vaswani2017attention}. For each of the multi-head attention blocks, there are two heads each of size 32, and the outputs of the three multi-head attention blocks are concatenated together.
The whole process is repeated three times to yield the final encoded unit features for the three groups of units, which will act as the attention keys being used in the selected units and the target unit heads, and also be reduced into a single vector being fed into the LSTM.

\subsection{Attention-based Pooling}
To reduce the extracted unit feature vectors of each group into a single vector, a simple average-pooling is applied in AlphaStar. In a similar scenario, the max-pooling is utilized in OpenAI five. Both operators are non-trainable which potentially limits their expressive capability. Inspired by the attention-based multiple instance learning \cite{ilse2018attentionMIL}, we propose the trainable attention-based pooling based on the multi-head attention. More specifically, the extracted unit feature vectors are treated as the keys and values in a multi-head attention, and multiple trainable weight vectors are created as the queries. The outputs of the multi-head attention are then flattened to yield the reduced vector, with the dimension being defined by the number of query weight vectors, and also the number of head and head size of the multi-head attention. It is observed that this trainable reduce operator gives better performance in supervised learning.

\subsection{Conditional Structures}
With the observations encoded into vector representations, concatenated and processed by a residual LSTM block, the action is then decided based on the LSTM output. To manage the structured, combinatorial action space, AlphaStar uses an auto-regressive action structure in which each subsequent action head conditions on all previous ones, via an additive auto-regressive embedding going through all heads. During the design of SCC, we found that it is critical for all other heads to condition on the selected action, which defines what to do, for example, moving, attacking, or training a unit, and for the two heads that decide the target unit or position, it is also helpful to condition on the selected units. However, for all other heads, it is not necessary to condition on each other, thus the order of these heads, say the skip frames and queued heads, also does not matter. In addition, in view of the limited expressive capability of the additive operator, we adopt the structure of concatenation followed by a fully connected layer, to provide full flexibility for the network to learn a better conditional relationship.

Due to the importance of the selected action for some action heads, we propose to condition on it further via the conditioned concat-attention. To be specific, the concat-attention is applied in the target unit head to output the probability distribution from which the target unit is sampled. In the original concat-attention \cite{luong2015effective}, the attention score is computed as follows:
\[\mathrm{score}(\mathbf{q},\mathbf{u}_i)=\mathbf{v}^T\mathrm{tanh}(\mathbf{W}[\mathbf{q};\mathbf{u}_i]),\]
where $\mathbf{q}$ is the query, $\mathbf{u}_i$ is the encoded unit feature vector serving as the key, $\mathbf{W}$ and $\mathbf{v}$ are the weights to be trained. In the proposed conditioned concat-attention, we replace the single weight vector $\mathbf{v}$ by the embedding of the sampled selected action. Since there are different embeddings for different selected actions, the conditioned concat-attention provides the capability of defining different function mappings for different selected actions. It makes sense intuitively, since different selected actions may need totally different criteria for selecting the target, for example, the repair action may want to target damaged alliance units and the attack action may want to target nearby enemy units. The same conditioned concat-attention is also applied inside the pointer network in the selected units head, in place of the original simple dot product attention, to also further condition on the selected action.

\section{Training Platform}\label{appendix-platform}
\begin{figure*}[ht]
    \centering
    \includegraphics[width=0.9\textwidth]{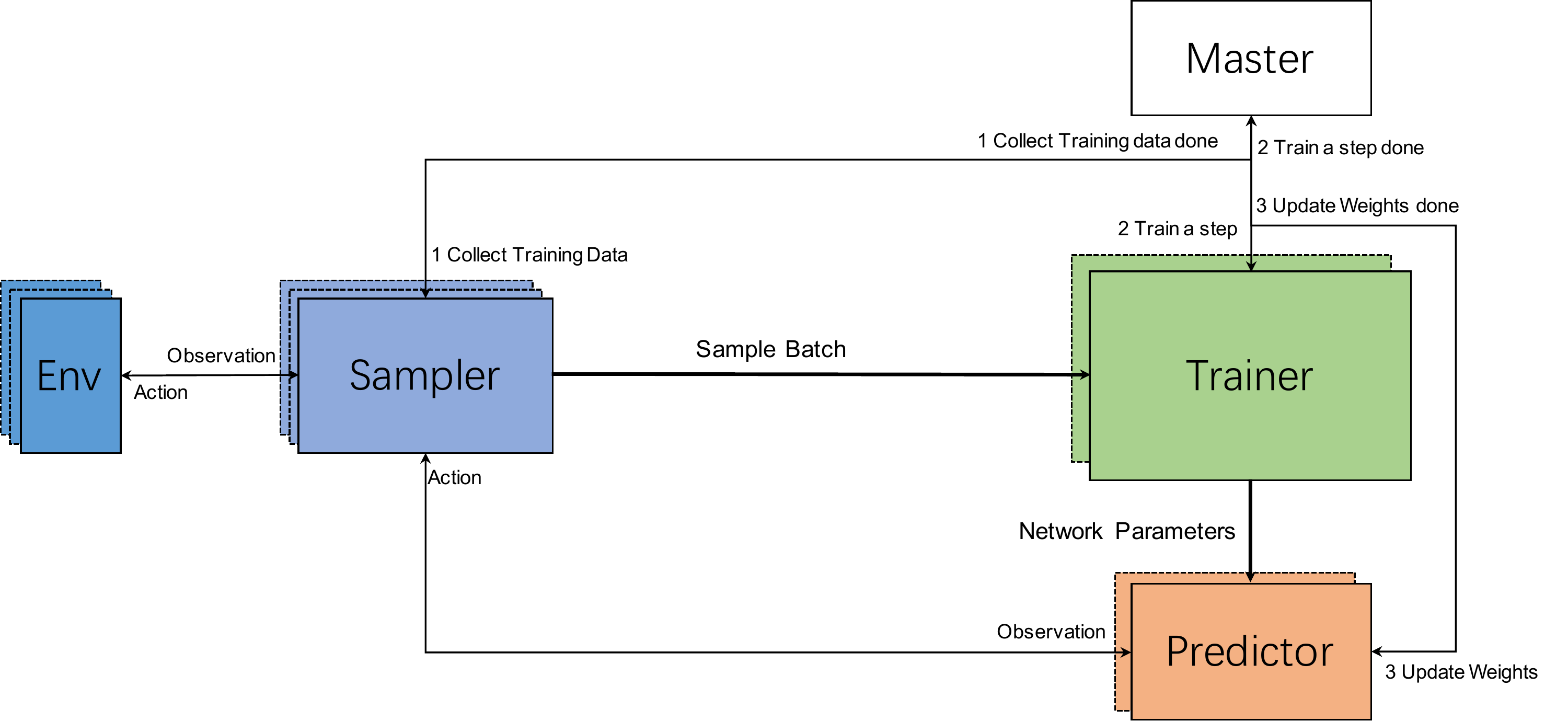}
    \caption{Diagram of the training process.}
    \label{fig:platform_train}
\end{figure*}

We developed a highly scalable training system based on the actor-learner architecture. The diagram of the training process for one agent is shown in Fig. \ref{fig:platform_train}. In our system, there are a number of Samplers, each of which continuously runs a single SC2 environment and collects training data. When a rollout of training data is generated, it will be sent to a Trainer over the network and saved into a local buffer. When there are sufficient training data in the local buffer, Trainer starts to execute a training step, during which it repeatedly samples a batch of training data, computes the gradient, and executes MPI all-reduce operation. When the training data has been utilized some number of times, this training step is terminated and the local buffer is cleared. After that, Trainers distribute updated network parameters to Predictors through network. Predictor provides batched inference service on GPU for Samplers to make efficient use of resource. For each agent, we run about 1000 (AlphaStar 16,000) concurrent SC2 matches, and can collect a sample batch of total size 144,000 and perform a training step in about 180 seconds. So about 800 (AlphaStar 50,000) agent steps can be processed per second on average.

\begin{figure*}[ht]
    \centering
    \includegraphics[width=0.9\textwidth]{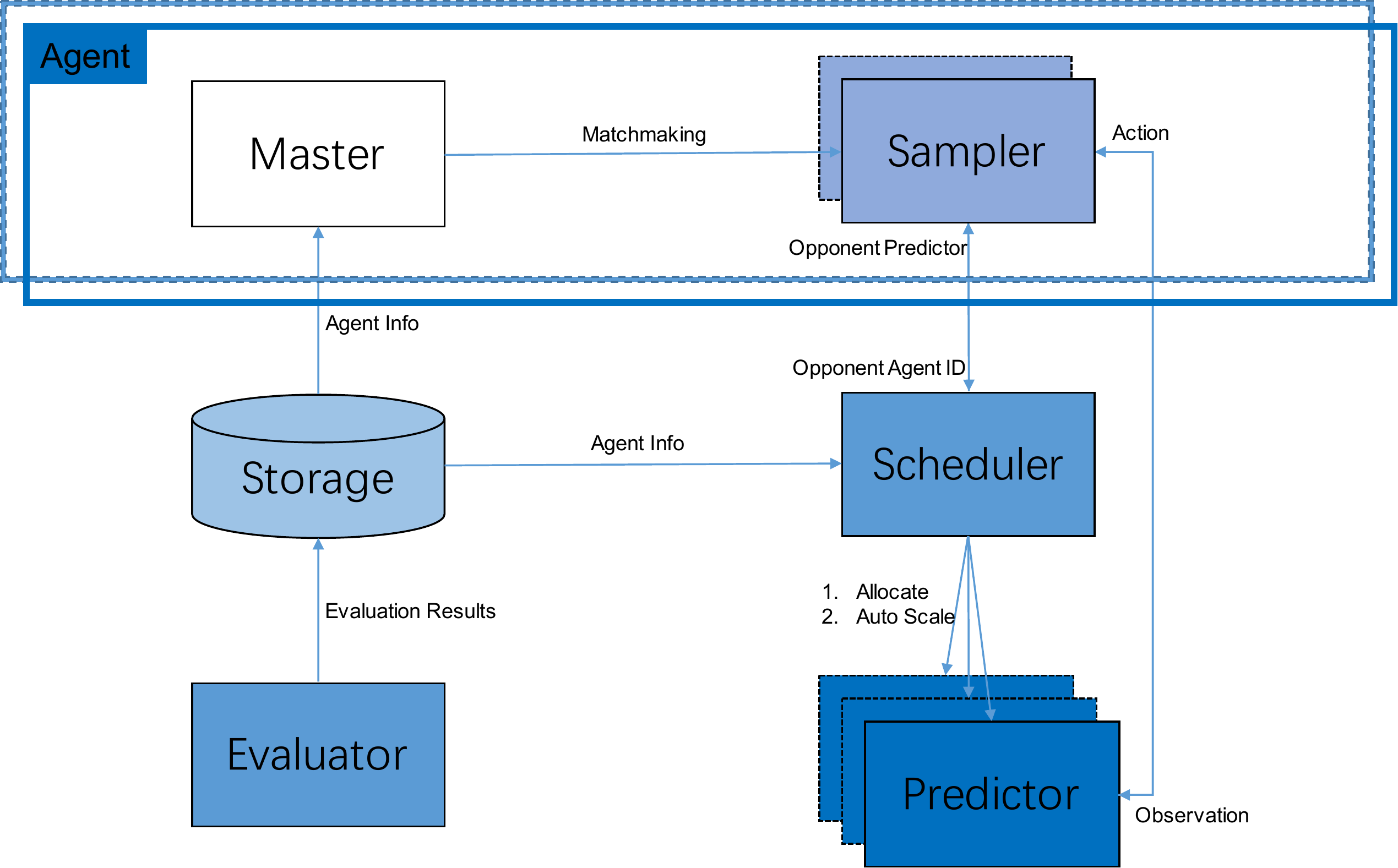}
    \caption{Diagram of the league training.}
    \label{fig:platform_league}
\end{figure*}

To support the whole league training, we also build a league system as shown in Fig. \ref{fig:platform_league}, which consists of four components, i.e., Storage, Predictor, Scheduler and Evaluator. We introduce each component in the following.
\begin{itemize}
    \item Storage: we use a MySQL DB service to store league information, like agent-id, agent-type, model-path and so on. The evaluation results of win rates between agents are also saved into the storage.
    \item Predictor: we use a single cluster of Predictors to provide inference service for all agents in the league, and the cluster is shared by training agents. Not only GPUs, CPUs are also used as Predictors. 
    \item Scheduler: Scheduler maintains the predictors and provides naming service, which receives agent-id and returns an available Predictor (allocate one if not exists). As the number of agents in league keeps growing, there is no guarantee for each agent with at least one GPU Predictor, in that case the CPU Predictors are used instead. The distribution of requests over agents also changes in the training process, Scheduler is also responsible to auto scale the Predictor number for agents according to the request amount.
    \item Evaluator: an Evaluator is needed to get the win rates between agents in league. The evaluation results are saved into Storage and will be used to calculate the matchmaking distribution.
\end{itemize}

\end{document}